\documentclass{article}

\usepackage{libertine} %
\usepackage[letterpaper,margin=1in]{geometry}

\usepackage{todonotes}
\usepackage{graphicx}

\usepackage[utf8]{inputenc} %
\usepackage[T1]{fontenc}    %
\usepackage{hyperref}       %
\usepackage{url}            %
\usepackage{booktabs}       %
\usepackage{amsfonts}       %
\usepackage{nicefrac}       %
\usepackage{microtype}      %
\usepackage{xcolor}         %
\usepackage{natbib}
\usepackage{amsmath}
\usepackage{amssymb}
\usepackage{amsthm}
\usepackage{subcaption}

\usepackage{todonotes}

\usepackage[space]{grffile}

\newtheorem{claim}{Claim}

\definecolor{forestgreen}{rgb}{0.0, 0.27, 0.13}

\title{The Calibration Generalization Gap\thanks{A version of this work appeared at the ICML 2022 Workshop on Distribution-Free Uncertainty Quantification.}}

\author{%
  A. Michael Carrell \\
  University of Cambridge\\
  \texttt{\sf ac2411@cam.ac.uk}
  \and
  Neil Mallinar \\
  UC San Diego \\
  \texttt{\sf nmallina@ucsd.edu}
  \and
  James Lucas \\
  NVIDIA \\
  \texttt{\sf jlucas@cs.toronto.edu}
  \and
  Preetum Nakkiran\\
  Apple \\
  \texttt{\sf preetum@apple.com}
}
\date{}

\newif\ifcomments
\commentstrue

\newcommand{\cX}{\mathcal{X}}
\newcommand{\cD}{\mathcal{D}}

\newcommand{\E}{\mathop{\mathbb{E}}}

\begin{document}

\maketitle

\begin{abstract}
Calibration is a fundamental property of a good predictive model:
it requires that the model predicts correctly in proportion to its confidence.
Modern neural networks, however, provide no strong guarantees on their calibration---and can be either poorly calibrated or well-calibrated depending on the setting.
It is currently unclear which factors contribute to good calibration (architecture, data augmentation, overparameterization, etc),
though various claims exist in the literature.

We propose a systematic way to study the calibration error:
by decomposing it into (1) calibration error on the \emph{train set}, and
(2) the \emph{calibration generalization gap}.
This mirrors the fundamental decomposition of generalization.
We then investigate each of these terms, and give empirical evidence that (1) DNNs are typically always calibrated on their train set,
and (2) the calibration generalization gap is upper-bounded by the standard generalization gap.
Taken together, this implies that \emph{models with small generalization gap (|Test Error - Train Error|) are well-calibrated}.
This perspective unifies many results in the literature, and suggests that interventions which reduce the generalization gap
(such as adding data, using heavy augmentation, or smaller model size) also improve calibration.
We thus hope our initial study lays the groundwork for a more systematic and comprehensive understanding of the relation between
calibration, generalization, and optimization.

\end{abstract}

\section{Introduction}
\label{sec:introduction}
When machine learning models are deployed in the real world, as components of larger systems,
we often want to know more about their behavior than just overall loss or accuracy.
For classification models, for example, it is helpful to know not only their test error, but also estimates of
predictive uncertainty --- how confident the model is on various inputs.
Calibrated models can be more useful than uncalibrated ones, because their confidences are operationally meaningful:
conditioning on high-confidence predictions is ``almost as good'' as conditioning on
high-accuracy predictions\footnote{Formally, this is true in expectation, for first-order moments.},
but can be done without knowing the ground-truth.

Many machine learning models produce outputs in a probability simplex,
and it is natural to ask whether these outputs are intrinsically calibrated
(even without post-hoc modifications).
For deep neural networks (DNNs), we can ask, informally:
\begin{center}
\textit{
Are DNNs typically well-calibrated, on standard tasks?
}
\end{center}
This question has received much attention over the last several decades,
but the literature remains muddled:
early works found that networks were reasonably calibrated \citep{niculescu2005predicting},
then \citet{DBLP:conf/icml/GuoPSW17} found that ``modern'' networks (in 2017) were poorly calibrated,
and most recently \citet{minderer_revisiting_2021} argued that current networks (in 2021) are in fact calibrated.
The issue is complicated because the notion of ``deep neural network'' has itself evolved over time:
with different architectures, optimizers, and even different benchmark datasets on which calibration is measured.

\begin{figure}[h!]
  \centering
  \includegraphics[width=0.9\linewidth]{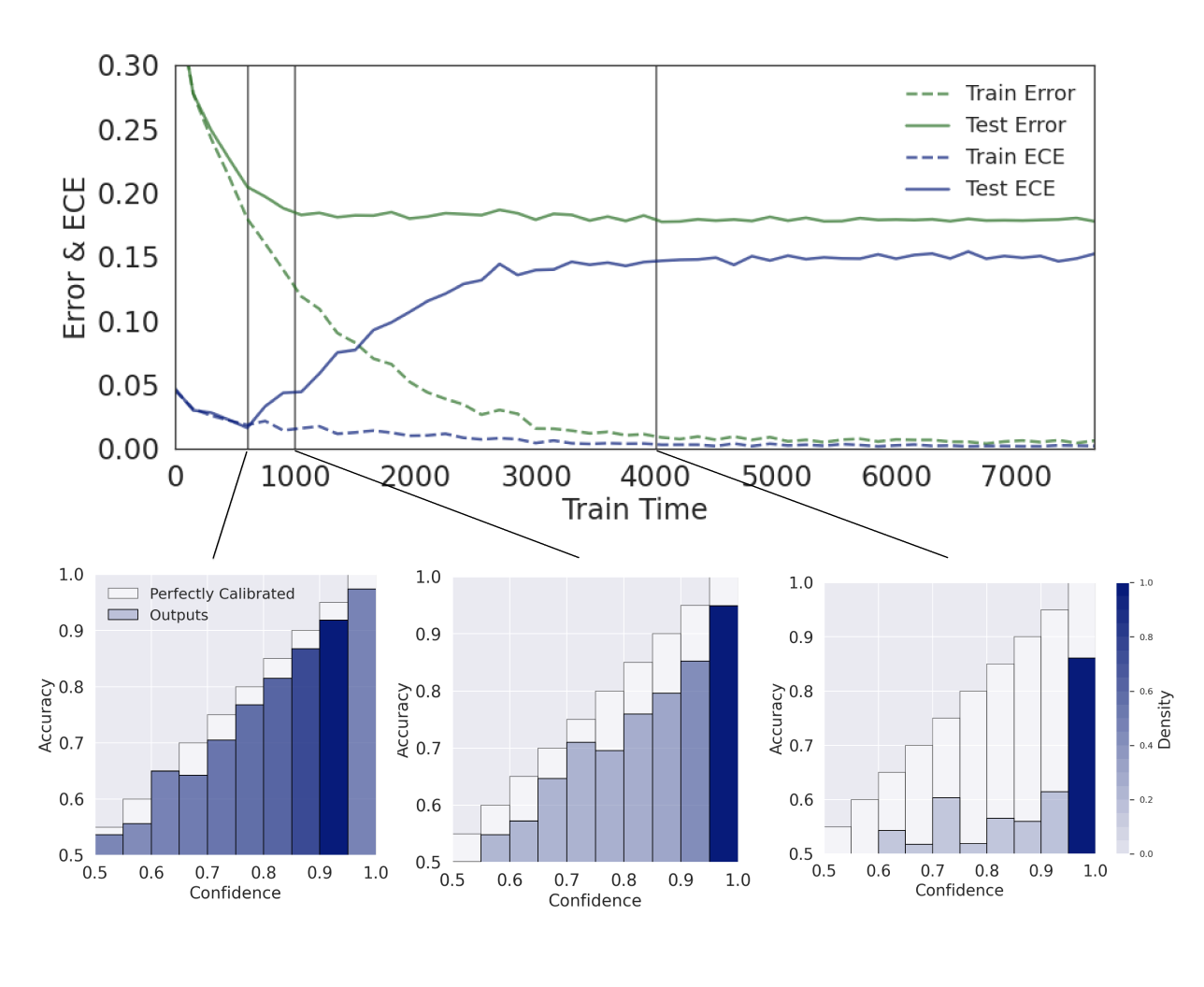}
  \caption{ECE and error of an overparameterized model throughout training. We show reliability diagrams at various point during training.
  In the early stage of training (before batch $1000$), both test and train ECEs are small ($< 0.05$), even though the test error itself is high ($> 20\%$).
  After batch $1000$, the generalization gap grows, and the test and train ECEs also diverge.
  In the late stage of training, as the model becomes overconfident, the test ECE approaches test error, while the train ECE goes to zero.
  Throughout, the difference in test and train ECE is upper-bounded by the difference in test and train error.
  }
  \label{fig:ece_time}
\end{figure}

A more refined question is thus to ask: \emph{when} are DNNs calibrated? 
That is, for what settings of architecture, optimizer, data distributions, etc.
Various prior works have argued that certain individual factors have significant impact of calibration---
for example, \citet{minderer_revisiting_2021} claims non-convolutional models tend to be better calibrated, and \citet{wen2020combining}
highlights the difference in calibration when using data augmentations.
A priori, a complete understanding of this question may require jointly understanding all of these factors---
but we can hope for better abstractions which let us study only
those aspects of our design choices which affect calibration.

In this work, we take steps to clarify the landscape of calibration in deep neural networks.
First, we propose a simple framework for reasoning about calibration:
Just as we can classically decompose the Test Error into the Train Error and the Generalization Gap,
we can similarly decompose the Test Calibration Error (Test ECE) as:
\begin{equation}
\label{eqn:gen-gap}
\underbrace{\textrm{TestECE}}_{\textrm{Calibration on Test Set}}
\leq
\underbrace{\textrm{TrainECE}}_{\textrm{Calibration on Train Set}}
+\ 
\underbrace{\left|\textrm{TestECE} - \textrm{TrainECE} \right|}_{\textrm{Calibration Generalization Gap}}
\end{equation}
This bounds the test calibration error in terms of an \emph{optimization} quantity (calibration on the train set) and 
a \emph{generalization} quantity.
We can then study these two terms individually, much like the classical approach of
generalization theory.

With this framework in hand, we then apply it to study the calibration of DNNs.
We first observe that the Train Calibration Error (TrainECE) is empirically \emph{uniformly close to 0},
even across different design choices (architectures, model size, train time, sample size, etc).
Notably, this holds even when the Train Error itself is still high: even DNNs which are poorly optimized
on their train set are still well calibrated on this train set.

\begin{claim}[Train Calibration, informal]
In most supervised learning settings, DNNs trained with cross-entropy loss are well-calibrated on their train sets.
That is,
\begin{equation}
    \textrm{TrainECE} \approx 0
\end{equation}
\end{claim}

We state this claim informally, since we do not yet understand how to precisely define ``most'' DNN settings,
and how to precisely quantify ``close to 0.''
Although this claim is not true universally, it appears to empirically hold across a variety of natural settings,
as we demonstrate in the experimental section\footnote{Understanding exactly the scope of this claim is an important open question.}. In particular, it empirically holds for both underparameterized and overparameterized models.

Now, assuming Claim 1, the TestECE is dominated by the \emph{calibration generalization gap}:
the difference in calibration
between train and test sets.
This leads us to study which factors affect the calibration generalization gap.
We find that the calibration gap is empirically upper-bounded by the standard generalization gap: the difference between test and train \emph{error}.

\begin{claim}[Calibration Generalization Bound, informal]
In most supervised learning settings, when trained with cross-entropy loss,
the calibration generalization gap is upper-bounded by the standard generalization gap:
\begin{equation}
\underbrace{\left|\textrm{TestECE} - \textrm{TrainECE} \right|}_{\textrm{Calibration Generalization Gap}}
\leq
\underbrace{\left|\textrm{TestError} - \textrm{TrainError} \right|}_{\textrm{Error Generalization Gap}}
\end{equation}
\end{claim}
This bound may not be surprising, since similar mechanisms are likely responsible for both kinds of generalization.
This claim is also not universally true, but appears to hold empirically in natural settings,
and we leave understanding its exact scope to future work.

Combining Claims 1 and 2, we find that the TestECE is upper-bounded by the generalization gap. Thus,
\emph{models with small generalization gap will have small test calibration error.}
In particular, underparameterized models will have small Test ECE,
while interpolating models, which fit their train sets, have large Test ECE
\footnote{
In fact, it is easy to see that overparameterized networks trained with cross-entropy loss will
never be perfectly calibrated, in the limit of infinite train time:
their Test ECE will converge to their Test Error.}.
This can be seen even by following a single large model throughout the course of its training, 
as it transitions from having a small generalization gap (early in training) to a large one 
(late in training, as it interpolates).
Figure \ref{fig:ece_time} illustrates training a ViT on CIFAR-10: the model is well-calibrated
in the early stages of training, but once the train error starts to diverge from the test error,
the model becomes overconfident and calibration starts to degrade.
Our results thus highlight the central role of \emph{generalization gap} in 
determining the calibration of DNNs.
Many prior results can be unified through this lens: interventions which decrease the generalization gap 
(like more data, stronger augmentation, stronger regularization, etc)
also tend to improve calibration.

\paragraph{Our Contributions}
\begin{enumerate}
    \item We propose the ``generalization decomposition'' of calibration, writing Test Calibration Error
    in terms of \emph{optimization} and \emph{generalization} components (Equation~\ref{eqn:gen-gap}).
    
    \item On the optimization side, we identify an apparently new optimization phenomenon:
    that supervised DNNs are often very well-calibrated on their train sets, throughout the entire
    trajectory of training (Claim 1).

    \item On the generalization side, we find that the calibration generalization gap
    is empirically upper-bounded by the standard generalization gap (Claim 2).
    This unifies many prior results in the calibration literature through the common lens of generalization gap.
    In particular, it predicts that
\begin{center}
\textit{
Models with small generalization gap will have small test calibration error.
}
\end{center}

\end{enumerate}
Our results clarify the existing literature on if and when DNNs are calibrated, by emphasizing the central
role of \emph{generalization gap}.
They also suggest important open problems for future work: to better quantify
and theoretically understand the empirical behaviors of Claims 1 and 2.

\section{Related Work}
\label{sec:related}

\paragraph{Calibration Measures.}
Brier scores \citep{brier1950verification} are often used in classification settings, measuring Mean Squared Error of a model's output probabilities against the ground truth class labels. 
\citet{naeini2015obtaining} propose to study discretized measures of calibration error using confidence bin groupings.
They study binned expected calibration error (binned-ECE), which operates by averaging the difference between prediction probabilities and predicted accuracy over all confidence bins, and Maximum Calibration Error (MCE) which reports the maximum difference over all bins.

Recent works have argued that ECE is not an ideal metric for calibration as it only considers the probability of the predicted class per datapoint, rather than all output probabilities.
\citet{nixon2020measuring} propose two calibration measures to resolve this issue, which they call Adaptive Calibration Error (ACE) and Static Calibration Error (SCE).
ACE and SCE extend ECE by measuring calibration over all classes in each bin, rather than just the predicted class. 

\paragraph{Recalibration.}
One way to address miscalibration is by \textit{recalibration} techniques.
There are two primary categories of recalibration that arise in practice, those that operate post-hoc, after training a model, and those that influence the optimization process itself.
Popular techniques for recalibration include Platt scaling \citep{platt1999probabilistic} and its more recent variant, temperature scaling \citep{DBLP:conf/icml/GuoPSW17}.
Additionally, it has been noted that calibration behaves differently in deep ensembles \citep{DBLP:conf/nips/Lakshminarayanan17, wen2020combining}, and when using data augmentations \citep{thulasidasan2019mixup, wen2020combining}, or label smoothing~\citep{szegedy2016rethinking}, for example.

\paragraph{Calibration in (Deep) Neural Networks.}
\citet{DBLP:conf/icml/GuoPSW17,DBLP:journals/corr/GuoPSW17} study ECE and MCE on modern (convolutional) neural networks 
and observed that neural networks are, a priori, miscalibrated. \citet{minderer_revisiting_2021} observed that recent model families, such as ViTs \citep{dosovitskiy2020image} and MLP-Mixers \citep{tolstikhin2021mixer} tend to be better calibrated than their earlier, convolutional, counterparts.
The authors provide suggestive evidence that a deeper understanding of model accuracy and intrinsic properties of architectures with relation to calibration is necessary to clarify the landscape of calibration in neural networks.\\

\noindent In this work, we study Expected Calibration Error as it is the primary metric studied in the
DNNs literature.
We are concerned with clarifying the \textit{instrinsic} properties of deep neural networks that contribute to their (mis)calibration, and determining whether such properties are affected by architecture.
As such, we do not consider recalibration techniques but rather study calibration with respect to general properties of neural networks such as their optimization and generalization trajectories.

\section{Preliminaries}
\label{sec:prelims}

We consider the setting of binary classification.
Let $\cD$ be a joint distribution over inputs $x \in \cX$ and labels $y \in \{0, 1\}$.
We consider models $f: \mathcal{X} \to [0, 1]$,
which map inputs $x \in \mathcal{X}$ to values in $[0, 1]$, which should
be interpreted as estimated probabilities $\Pr[y=1 | x]$.
In the binary classification setting, a model $f$ is
\emph{perfectly calibrated with respect to distribution $\cD$} \citep{brocker2009reliability} if the following holds.
\begin{equation}
    \forall \ell \in [0, 1]: \quad \mathop{\E}_{x, y \sim \cD}[y \mid f(x) = \ell] = \ell
\end{equation}

To measure degree of miscalibration, we use the \emph{Expected Calibration Error (ECE)},
defined with respect to a distribution $\cD$ as:
\begin{equation}
    \label{eqn:true_calibration_error}
    \textrm{ECE}_{\cD}(f) := \E_{\cD}[ |\E_{\cD} [y \mid f(x)] - f(x)| ]
\end{equation}
We estimate the ECE using histogram binning, as is standard in the literature \citep{naeini2015obtaining,DBLP:conf/icml/GuoPSW17}.
Note that the ECE and the binning-estimator have well-known issues (e.g. discontinuity, and sensitivity to binning hyperparameters) \citep{minderer_revisiting_2021,lee2022t,DBLP:conf/nips/KumarLM19}.
However, we use the ECE as a first step in studying calibration,
and we leave the study of smoother metrics to future works.

Equation \eqref{eqn:true_calibration_error} allows us to define calibration on both the train set and the test distribution. The \emph{TestECE}, which is the standard object of study, is simply $\textrm{ECE}_\cD(f)$
where $\cD$ is the test distribution.
Analogously, the \emph{TrainECE} is defined as $\textrm{ECE}_{\hat{\cD}}(f)$,
where $\hat{\cD}$ is the empirical distribution of train samples $\{(x_i, y_i)\}$.

\section{Experiments and Analysis}
\label{sec:calibration_gap}

\subsection{Experimental Setup}\label{sec:setup}
\paragraph{Dataset: CIFAR-5m binary.}
We perform experiments on the CIFAR-5m dataset introduced by \citet{nakkiran2021bootstrap}.
CIFAR-5m is a synthetic dataset constructed by sampling $32 \times 32$ images from a generative model trained on CIFAR-10 \citep{cifar}.
Samples are then labeled with a large transfer model \citep{ho_denoising_2020,DBLP:conf/eccv/KolesnikovBZPYG20}.
In total there are 5 million training data samples and 1 million test samples in CIFAR-5m, over the same ten classes in CIFAR-10.
We use the first 10k test samples for our held-out evaluations.
\citet{nakkiran2021bootstrap} observe that models trained on CIFAR-5m perform comparably on the CIFAR-10 test set as compared to models trained directly on the CIFAR-10 training data.
This suggests that CIFAR-5m has wider use as a larger proxy for CIFAR-10.
The experiments presented in this section use a binary version of CIFAR-5m comprising of the ``car'' and ``truck'' classes.
We present 10-class versions of these experiments in Appendix \ref{apdx:multiclass_results}.

\begin{figure}[h!]
\centering
\begin{subfigure}{0.4\textwidth}
    \includegraphics[width=\textwidth]{{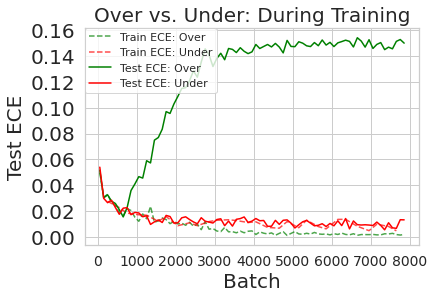}}
    \caption{During training, 10k samples.}
    \label{fig:during_over_under}
\end{subfigure}
\hspace{1cm}
\begin{subfigure}{0.4\textwidth}
    \includegraphics[width=\textwidth]{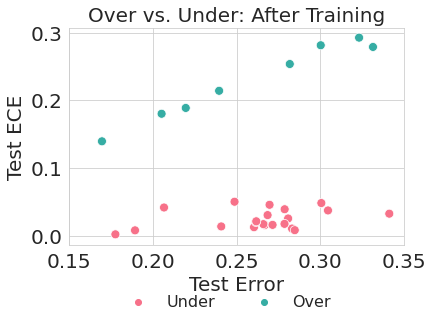}
    \caption{After training, varying samples.}
    \label{fig:after_over_under}
\end{subfigure}
\caption{Final test error vs. test expected calibration error, over and under parameterized models. The under parameterized models display low ECE while the validation error increases. However, the ECE of over parameterized models increases alongside validation error. 
}\label{fig:over_under}
\end{figure}

\paragraph{Models.} 
For single-model experiments (Figures \ref{fig:ece_time}, \ref{fig:over_under}, and \ref{fig:apdx_overunder}) we use the small variant of the Vision Transformer (ViT) \citep{dosovitskiy2021an} with patch size of $4 \times 4$ due to the size of images in CIFAR-5m.
For experiments considering multiple architectures (Figures \ref{fig:gap} and \ref{fig:apdx_gap}), we study a range of convolutional and non-convolutional architectures, to have a comprehensive view of calibration across model families. 
They include: preactivation ResNet \citep{he2016identity}, ViT, MLP-Mixer \citep{tolstikhin2021mixer}, AlexNet \citep{krizhevsky2012imagenet}, DenseNet \citep{huang2017densely}, and vanilla (deep) CNNs.

\paragraph{Training.} 
In order to compare models, we train directly on CIFAR-5m Binary, without pretraining, so that we can study the effects of amount of parameterization seen in training. 
For the models which are effectively overparameterized, we train on a fixed-size subsample of $10$k images for $100$ epochs, unless otherwise stated. 
For the effectively underparameterized models, we train for $100$ epochs, but with fresh samples at each epoch (i.e. one epoch on 1 million training samples).
The preactivation ResNet, AlexNet, and DenseNet models are trained using SGD with learning rate 0.1, weight decay 0.0, and momentum 0.9. We additionally employ cosine annealing on the learning rate over 100 epochs.
ViTs, MLP Mixers, and and VGG models are trained using Adam with learning rate $10^{-4}$, and weight decay $10^{-5}$.
We do not decay the learning rate in this setting.

We train without data augmentations as the goal of our work is to compare overparameterized models to underparameterized models.
In the overparameterized case we are interested in models that can easily fit their training datasets so that they incur higher generalization error.
In the underparameterized case, we have 5 million training samples so we are able to sample new images in each epoch.
We choose hyperparameters for each model family based on performance on the CIFAR-10 test set. 
Evaluation is done on the held-out CIFAR-5m test set.
For details, see the appendix.

\subsection{Results}

We use the calibration generalization gap to study calibration behavior in both under and overparameterized regimes. We start with analyzing the calibration generalization gap in the overparameterized regime, to clarify how the gap evolves during training. We then look at the difference between the overparameterized and underparameterized regimes. Finally, we compare the calibration generalization gap to the standard error generalization gap, finding the error generalization gap acts as an upper-bound for the calibration generalization gap.

\paragraph{Calibration gap of overparameterized models.} Figure \ref{fig:ece_time} shows the train error, test error, and ECE of an overparameterized model throughout the course of training. The solid lines correspond to test values, while dashed lines correspond to train values. The vertical lines correspond to the points at which we compute the calibration values displayed in the reliability diagrams. The reliability diagrams capture the calibration of the model at various points in training. Our reliability diagrams deviate from the standard, combining the reliability diagram and the confidence histograms often seen as a pair. The clear bars reflect perfectly calibrated accuracy for a given bin, and the height of the outputs reflects the model's accuracy in the given bin. The bars are shaded according to the density of confidences in the corresponding bin. 

We see the test ECE drops along with test error, until the first snapshot at batch $608$. The corresponding train values also decrease. We observe the reliability diagram at batch $608$. The test error here is still high, but the model is well-calibrated. The test error then continues to go down, but the test ECE starts to increase. Train values continue to go down.
At batch $992$, we look at the calibration again. We note the model starts to become overconfident here, placing most of its predictions at or above $0.9$. The test and train error and train ECE all continue to go down, but the test ECE continues to increase. By batch $4032$, all values have converged to their final values. Train error and ECE are approaching $0$, and test ECE is close to test error \footnote{We note that, in the limit, test ECE converges to test error for overparameterized models. The gap seen here could be due to a combination of factors, including the finite train time, nonzero learning rate at the end of training, and the effect of binning on calibration values.}. We inspect the calibration at batch $4032$ and find the model is poorly calibrated and overconfident, placing most of its predictions in the top bin.

Figure~\ref{fig:ece_time} highlights how the calibration generalization gap of overparameterized models evolves during training, with train ECE converging to $0$ and test ECE ending close to test error. Next, we explore the role that effective parameterization plays in calibration.

\paragraph{Over vs under parameterized: difference in calibration.} Figure \ref{fig:over_under} shows error vs. ECE for over and underparameterized models, after and during training. In Figure \ref{fig:after_over_under} we observe the final test error and ECE for overparameterized models, with varying sample sizes. We train $8$ overparameterized models for $100$ epochs on fixed-dataset sizes ranging from $500$ to $10,000$ samples. We also measure test error and ECE for one underparameterized model at similar points in test error during training\footnote{We do this for computational ease, but we find similar results apply when taking test error and ECE values only at the end of training in the underparameterized case. This is due to the test ECE staying low throughout training for underparameterized models.}. 

We see the test error of underparameterized models stays low, consistently at or below $0.05$, for a large test error range. We find this does not hold for overparameterized models: as test error increases, so does ECE. We note the linear correlation between final test error and ECE for overparameterized models. The results show a large difference between the final calibration of over and underparameterized models.

Figure \ref{fig:during_over_under} shows ECE during training for an over and underparameterized model. The train ECE for the over and underparameterized model both go to zero, as well as the underparameterized model's test ECE. The overparameterized model's test ECE stays close to the underparameterized model's, until about batch $500$. Then the overparameterized model's test ECE increases until converging. We note the over and underparameterized models have diverging calibration generalization gaps at the end of training as a result. The results show the difference between over and underparameterized models' test ECE emerges early in training and leads to overparameterized models having a large calibration generalization gap, while underparameterized models have a gap close to zero.

\begin{figure}[h!]
\centering
\begin{subfigure}{0.4\textwidth}
    \includegraphics[width=\textwidth]{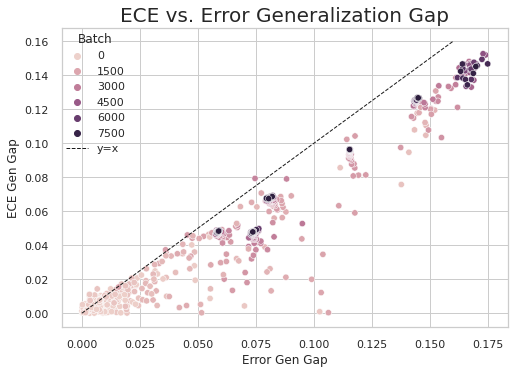}
    \caption{All architectures during training.}
    \label{fig:gap_during}
\end{subfigure}
\hspace{1cm}
\begin{subfigure}{0.4\textwidth}
    \includegraphics[width=\textwidth]{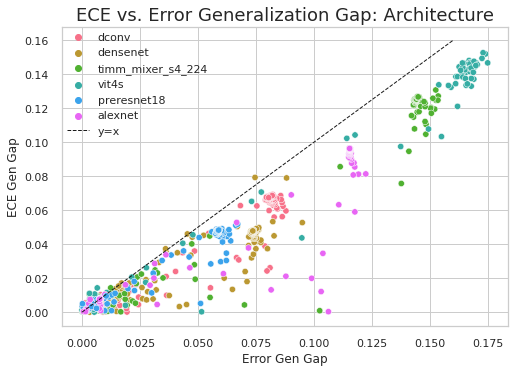}
    \caption{Throughout training, by architecture.}
    \label{fig:gap_all}
\end{subfigure}
\caption{ECE generalization gap vs error generalization gap, varying architectures. \textit{(left)} Color corresponds to batch, with lighter colors being earlier in training. \textit{(right)} Color corresponds to model family. Note there is a linear relationship between error generalization gap and ECE gap throughout training, with the former upper bounding the latter.
}
\label{fig:gap}
\end{figure}

\paragraph{Error Generalization Gap vs. Calibration Generalization Gap.} We now give empirical evidence for Claim 2. Figure \ref{fig:gap} shows the error generalization gap versus the calibration generalization gap of overparameterized models,
which interpolate at the end of training.
We train models of various architectures in the setup described in Section \ref{sec:setup}, and measure their error and calibration generalization gaps throughout training. Figure \ref{fig:gap_during} shows the results stratified by training batch: the calibration generalization is upper bounded by the error generalization gap, throughout training.
Figure \ref{fig:gap_all} shows the same experiments stratified by architecture, showing that the bound holds across 
many diverse architectures.

\section{Conclusion}
In this work we proposed studying the ``generalization decomposition'' of calibration,
which decomposes calibration error
into optimization and generalization terms.
We gave empirical evidence that supervised DNNs almost always achieve close to zero train calibration error (Claim 1),
and that the calibration generalization gap is upper-bounded by the error generalization gap (Claim 2).
Taken together, these claims imply that 
\emph{models with small generalization gap will have small test calibration error}.
This unifies the many disparate factors which are known to influence calibration (architecture, optimizer, etc)
via the common lens of generalization.
We hope our initial study inspires future work to understand
the interplay between calibration, generalization, and optimization in machine learning.

\subsubsection*{Author Contributions}
AC designed and ran most of the experiments, wrote the initial draft of the paper and contributed to revisions, and led the team.
NM helped in development of experiments and code, and conducted multi-class experiments. 
JL advised on the project and contributed to elements of experiment design and analysis.
PN managed and advised the project, and contributed to the theoretical formulation and experiment design.
All authors contributed to writing the paper.

\subsubsection*{Acknowledgements}
We thank Anastasios Angelopoulos for their
constructive reviewing feedback on our ICML workshop submission,
and Russ Webb for comments on a later draft. 

This work in supported in part by an EPSRC i-CASE Fellowship, with the corporate sponsor Microsoft Research.

\label{sec:conclusion}

\bibliographystyle{plainnat}
\bibliography{refs}

\newpage
\appendix
\section{Multi-class Results}
\label{apdx:multiclass_results}

\begin{figure}[h!]
\centering
\begin{subfigure}{0.45\textwidth}
    \includegraphics[width=\textwidth]{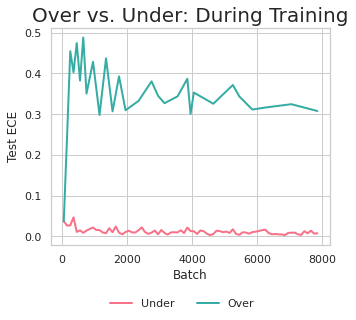}
    \caption{During training, 25k samples.}
\end{subfigure}
\hspace{1cm}
\begin{subfigure}{0.45\textwidth}
    \includegraphics[width=\textwidth]{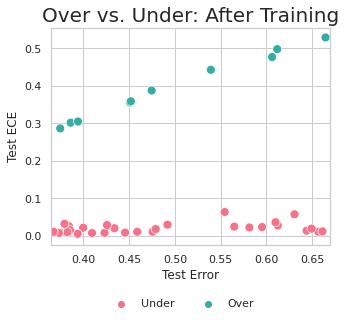}
    \caption{After training, varying samples.}
\end{subfigure}
\caption{We compare vit4s models trained on 10 class CIFAR-5m in both the overparameterized and underparameterized settings. \textit{(left)} We plot test ECE vs. test error in the overparameterized setting over numerous sample sizes, showing a near-linear trend between test ECE and test error. \textit{(right)} We plot test ECE vs. batch number during the training of an overparameterized model ($n=25k$) and it's equivalent underparameterized counterpart.}
\label{fig:apdx_overunder}
\end{figure}

\begin{figure}[h!]
\centering
\begin{subfigure}{0.4\textwidth}
    \includegraphics[width=\textwidth]{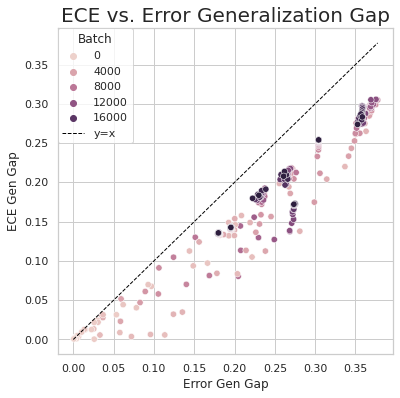}
    \caption{All architectures during training.}
\end{subfigure}
\hspace{1cm}
\begin{subfigure}{0.45\textwidth}
    \includegraphics[width=\textwidth]{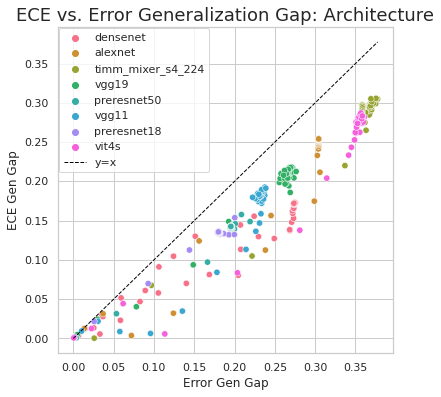}
    \caption{Throughout training, by architecture.}
\end{subfigure}
\caption{We plot ECE generalization gap vs. error generalization gap for various deep neural network model classes and model sizes when trained on 10 class CIFAR-5m, $n=25k$. \textit{(left)} We color-code points by the batch number during training, with lighter colors representing the early stages of training and darker colors for the late stages of training. \textit{(right)} We color-code points by model family and size, showing that observed trends hold over both architecture and model size.}
\label{fig:apdx_gap}
\end{figure}

\end{document}